\documentclass[runningheads]{llncs}
\usepackage{subfig}
\usepackage{graphicx}
\usepackage{wrapfig}

\usepackage{amsmath,amssymb} 
\usepackage{color}
\usepackage[width=122mm,left=12mm,paperwidth=146mm,height=193mm,top=12mm,paperheight=217mm]{geometry}
\newsavebox{\bigleftbox}
\usepackage{multirow}
\usepackage{hhline}
\usepackage{hyperref}
\usepackage{subfig}
\usepackage[table,xcdraw]{xcolor}
\begin{document}

\title{Characterization of Visual Object Representations in Rat Primary Visual Cortex}

\titlerunning{Visual Objects Representation in Rat's V1}
%
\author{Sebastiano Vascon \inst{1,2}$^{,}$\thanks{equal contribution} \orcidID{0000-0002-7855-1641} \and
Ylenia Parin \inst{1}$^{,\star}$ \and \\
Eis Annavini \inst{3}$^{,\star}$ \orcidID{0000-0002-3507-4873} \and
Mattia D'Andola \inst{3} \orcidID{0000-0003-4554-8822} \and \\
Davide Zoccolan \inst{3} \orcidID{0000-0001-7221-4188} \and \\
Marcello Pelillo \inst{1,2} \orcidID{0000-0001-8992-9243}
}
%
\authorrunning{Vascon et al.}

\institute{DAIS, Ca' Foscari University of Venice, Via Torino 155, 30170, Mestre (VE)\and
ECLT, Ca' Foscari University of Venice, San Marco 2940, 30124, Venice (VE) \and
Visual Neuroscience Lab, International School for Advanced Studies (SISSA), Via Bonomea 265, Trieste (TS)}
\maketitle              

\begin{abstract}
For most animal species, quick and reliable identification of visual objects is critical for survival. This applies also to rodents, which, in recent years, have become increasingly popular models of visual functions.
For this reason in this work we analyzed how various properties of visual objects are represented in rat primary visual cortex (V1). The analysis has been carried out through supervised (classification) and unsupervised (clustering) learning methods. We assessed quantitatively the discrimination capabilities of V1 neurons by demonstrating how photometric properties (luminosity and object position in the scene) can be derived directly from the neuronal responses. 
\keywords{rat's visual system, core object recognition, objects classification}
\end{abstract}

\section{Introduction}
For most animals, recognition of visual objects is of paramount importance. The visual system of many species has adapted to quickly and effortlessly detect and classify objects in spite of major variation (or transformation) in their appearance. This set of abilities is called Core Object Recognition \cite{DiCarloCox2007} and is typical of primate species, where a hierarchy of visual cortical areas, known as the ventral stream, supports shape processing and image understanding \cite{DiCarloZoccolanRust2012}. In recent years, some authors \cite{TafazoliEtAl2017,vermaercke2014functional} have investigated whether such core ability also exists in rats, by exploiting machine-learning tools, such as information theory and pattern classifiers, which have proved to be invaluable tools to understand how object vision works in primates ~\cite{DiCarloCox2007}. Indeed, rodents have become increasingly interesting model organisms to study the mammalian visual system \cite{zoccolan2015invariant,GlickfeldOlsen2017,glickfeld2014mouse,huberman2011can}. In particular, rodent object-processing abilities are supposed to be located along a progression of cortical areas, starting in primary visual cortex (V1), and extending to lateral extrastriate areas LM, LI, and LL~\cite{GlickfeldOlsen2017,SerenoAllman1991}, which are thought to be an homologous of the primate ventral stream. Recent work by \cite{TafazoliEtAl2017} has shown that, indeed, visual object representations along this progression become more explicit, i.e.: 1) information about low-level visual properties, such as luminance, is gradually lost; and 2) object identity becomes more easily readable through linear classifiers, even in the presence of changes in object appearance.

	\begin{figure}[t!]
    \vspace{-0.2cm}
    \footnotesize 
		\centering
        \sbox{\bigleftbox}{%
          \begin{minipage}[b]{.5\textwidth}
            \centering
            \vspace*{3mm}
            \subfloat[Objects tree structure\label{subfig-1:obj_tree}]
              {\includegraphics[width=\textwidth]{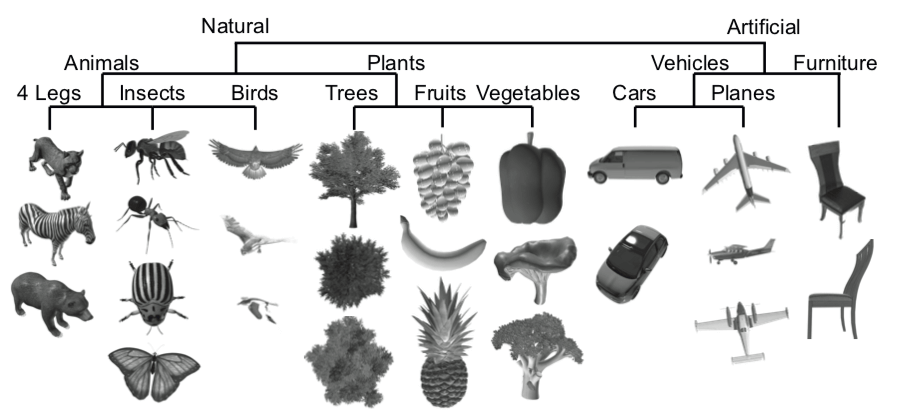}\label{fig:test1}}
          \end{minipage}%
        }\usebox{\bigleftbox}%
        \begin{minipage}[b][\ht\bigleftbox][s]{.5\textwidth}
        	\centering
            \subfloat[LH]
              {\includegraphics[width=0.3\textwidth]{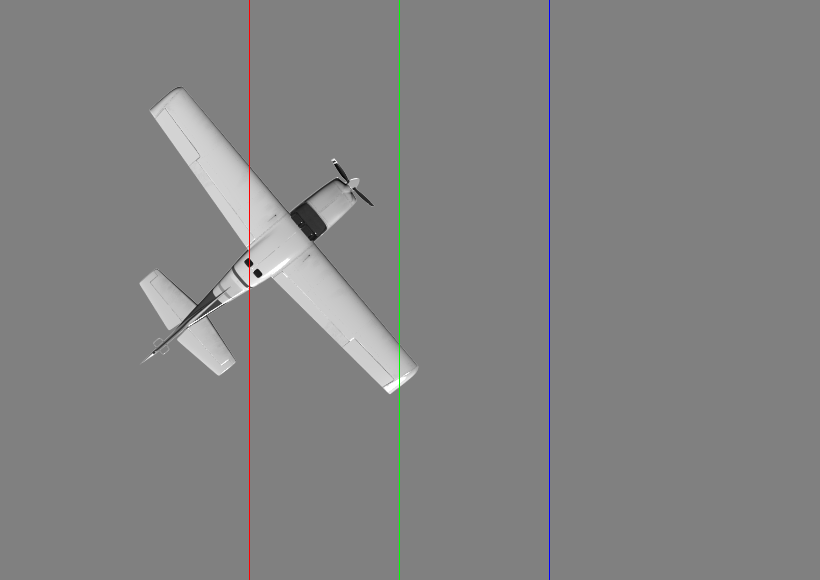}\label{fig:stimulusLH}}
            \hspace*{0.1mm}
              \subfloat[CH]
              {\includegraphics[width=0.3\textwidth]{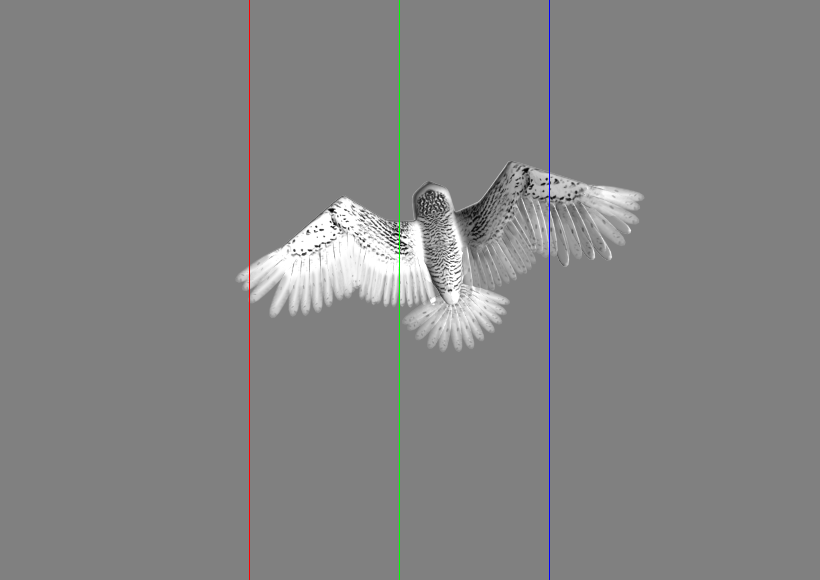}\label{fig:stmulusCH}}
            \hspace*{0.1mm}
            \subfloat[RH]
              {\includegraphics[width=0.3\textwidth]{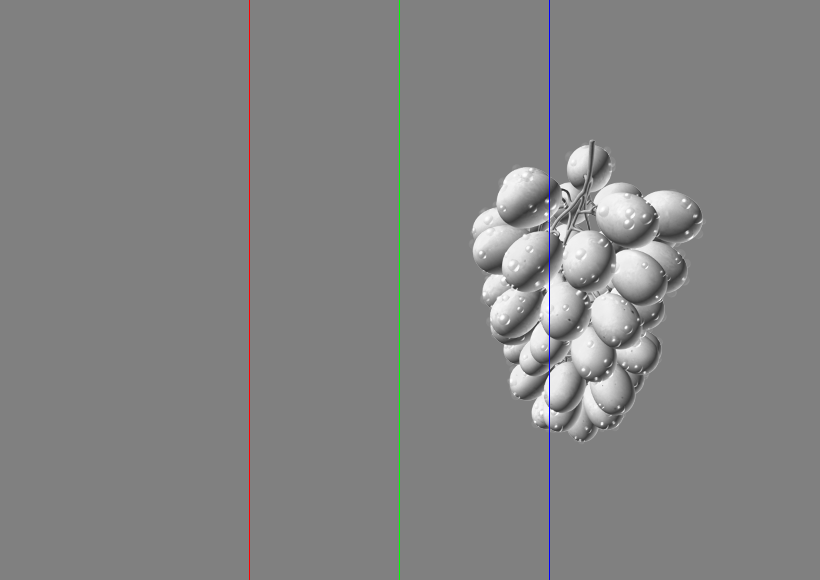}\label{fig:stmulusRH}}\\
            \subfloat[LL]
              {\includegraphics[width=0.3\textwidth]{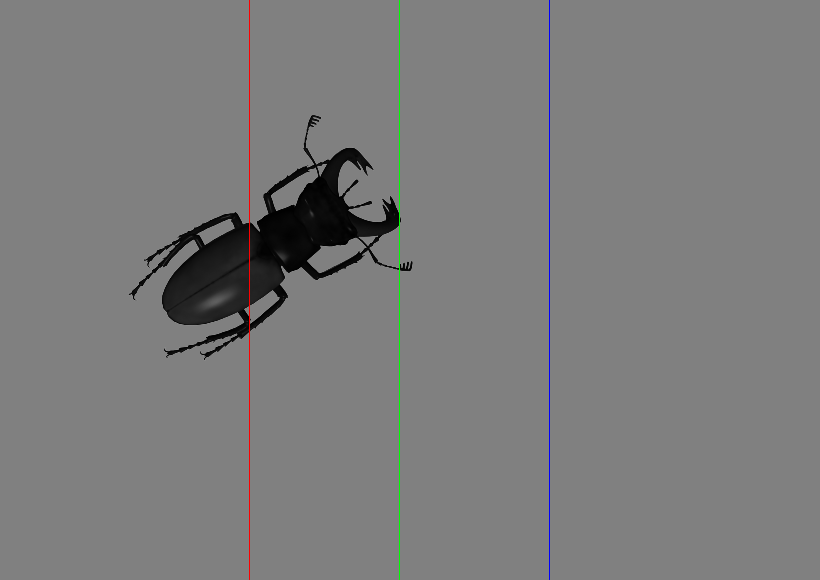}\label{fig:stmulusLL}}
            \hspace*{0.1mm}
            \subfloat[CL]
              {\includegraphics[width=0.3\textwidth]{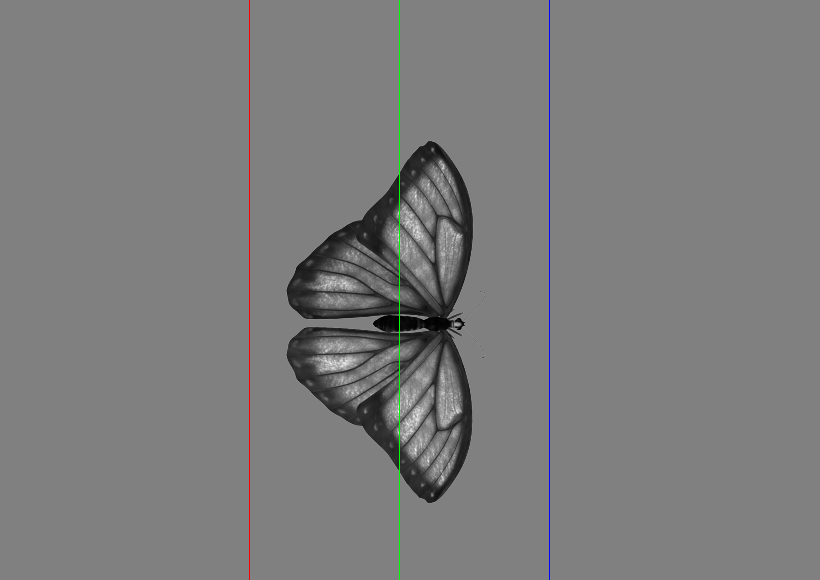}\label{fig:stmulusCL}}
            \hspace*{0.1mm}
            \subfloat[RL]
              {\includegraphics[width=0.3\textwidth]{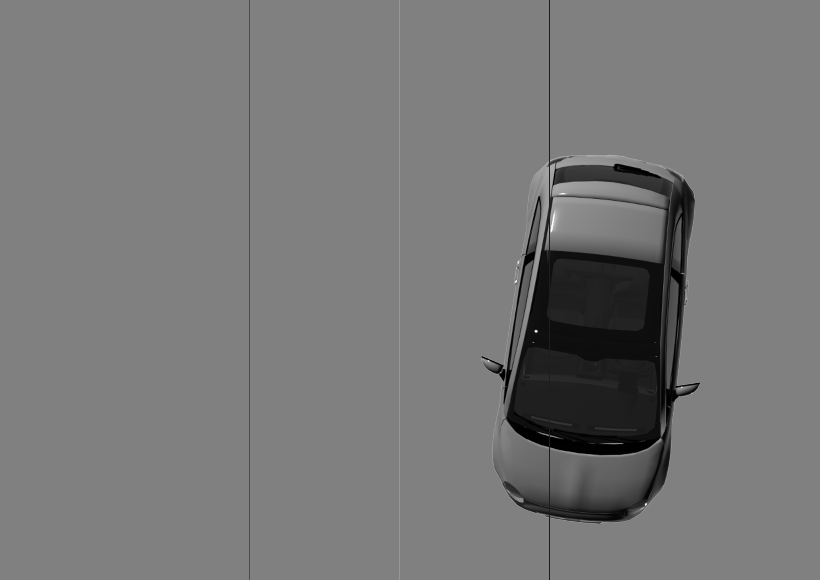}\label{fig:stmulusRL}}
        \end{minipage}
        \vspace*{3mm}
        \caption{\footnotesize \textbf{Stimulus set.} The image on the left shows the semantic hierarchical structure of the objects belonging to the stimulus set. The images on the right show an example stimulus for each combination of position (first letter of the label) and luminosity (the second letter), e.g., ``LH'' stands for left and high luminosity, ``RL'' stands for right and low luminosity, ``CL'' stands for center and low luminosity, etc.}
        \label{fig:stimulus_set}
        \vspace{-0.4cm}
	\end{figure}
    
In this work, we tried to understand at a deeper level what visual properties are encoded in the activity of a population of rat V1 neurons, using both unsupervised and supervised machine learning algorithms. The focus on V1 was motivated by the fact that this cortical area is the entry stage of visual information in cortex (future work will aim at providing a similar characterization in higher-order visual cortical areas). 

Specifically, in this study, we applied for the first time the Dominant Set\cite{pavan2007dominant} clustering algorithm (DS) to understand the structure of visual object representations in a visual cortical area.  The choice of the DS has been driven also by it recent success in related fields, like in brain connectomic \cite{Dodero2013,Dodero2015} or neuroscience \cite{Pennacchietti2017}, making it a good candidate for the task at hand. Furthermore, we applied an array of supervised algorithms to show that V1 neuronal responses can be used to predict with great accuracy the photometric information on the scene presented to the rat.

The article is organized as follows: in section \ref{sec:materials} we provide a description of the experimental methods that were used to produce the stimulus set and to record the responses of V1 neurons; in section \ref{sec:unsupervised_analysis} 
we describe the analysis we carried out to understand the organization of visual stimuli in terms of V1 neuron responses; in section \ref{sec:supervised_analysis} we show how V1 neuronal responses can be used to classify some key visual properties of the  stimuli, i.e., their location within the visual field and their luminosity; the section \ref{sec:conclusion} concludes the paper with some future perspectives. 
\section{Materials \& Methods}\label{sec:materials}
\vspace{-0.2cm}
In this section, we describe the steps that were performed to build the dataset and, whether non-conventional, the methodologies used to analyze the data.
\subsection{Stimulus set and data acquisition}
For our experiments, we built a rich and ecological stimulus set using a large number of objects, organized in a semantic hierarchy (Figure \ref{fig:stimulus_set}). 
To build the stimulus set we used 40 3D models of real world objects\footnote{\tiny TurboSquid \url{https://www.turbosquid.com/}}, both natural and artificial, each rendered in 36 different poses, randomly chosen around four main views (frontal, lateral, top, and 45$^{\circ}$ in azimuth and elevation), at one of three possible sizes (30-35-40$^{\circ}$) chosen at random, in one of three possible positions (0$^{\circ}$, $\pm 15^{\circ}$), also chosen at random, and rotated in plane of either 0, 90 or $\pm 45^{\circ}$ for a total of 1440 stimuli.
To further characterize the stimulus set we extracted a set of low and mid level features (such as position, contrast, and orientation) of the stimuli as they were presented on-screen to the rat: for the scope of the current work we will focus on the \emph{position} of the center of mass, and on the \emph{luminosity}. Stimuli were presented on a gray background to anesthetized na\"{i}ve Long-Evans rats for 150 ms while collecting extracellular neuronal activity from all the layers of primary visual cortex (V1) using multi-shank, 64-channel silicon electrode arrays\footnote{\tiny NeuroNexus Technologies, Ann Arbor, MI, USA}. We recorded extracellular potentials using an RZ2 BioAmp signal processor\footnote{\tiny Tucker-Davis Technologies, Alachua, FL, USA} at a sampling frequency of 24.4141 kHz. We characterized the neurons by carefully mapping the positions of each unit's receptive field (RF), rotating the rat afterwards in order to center the RFs on the screen and thus achieve maximal response to the stimuli.
\subsection{Data preprocessing}
We filtered the raw extracellular potentials with a band-pass filter (0.5--11 kHz) to extract neurons' spiking activity, and the resulting action potentials (spikes) were extracted using an Expectation-Maximization clustering algorithm \cite{rossant2016spike} that separates the spikes produced by different neurons according to their shape. Then, we estimated the optimal spike count window for each neuron using its firing rate averaged over the 10 best stimuli \cite{ZoccolanEtAl2013,TafazoliEtAl2017}, and we used it to compute the average number of spikes produced by a neuron in response to each stimulus, across its repeated presentations. Finally, we scaled the spike counts of each neuron to zero mean and unitary variance to obtain the population vectors for the stimulus set~\cite{KianiEtAl2007}. This led to a vector of size 177 for each visual stimulus that was used in all unsupervised and supervised analysis, where 177 is the total number of units (single- and multi-unit) obtained through spike sorting.

\begin{figure}[t!]

		\centering
        \sbox{\bigleftbox}{%
          \begin{minipage}[b]{.4\textwidth}
            \centering
              \vspace*{\fill}
              \subfloat[\footnotesize Pos hist.\label{subfig-histo-position}]{%
         \includegraphics[width=0.9\textwidth]{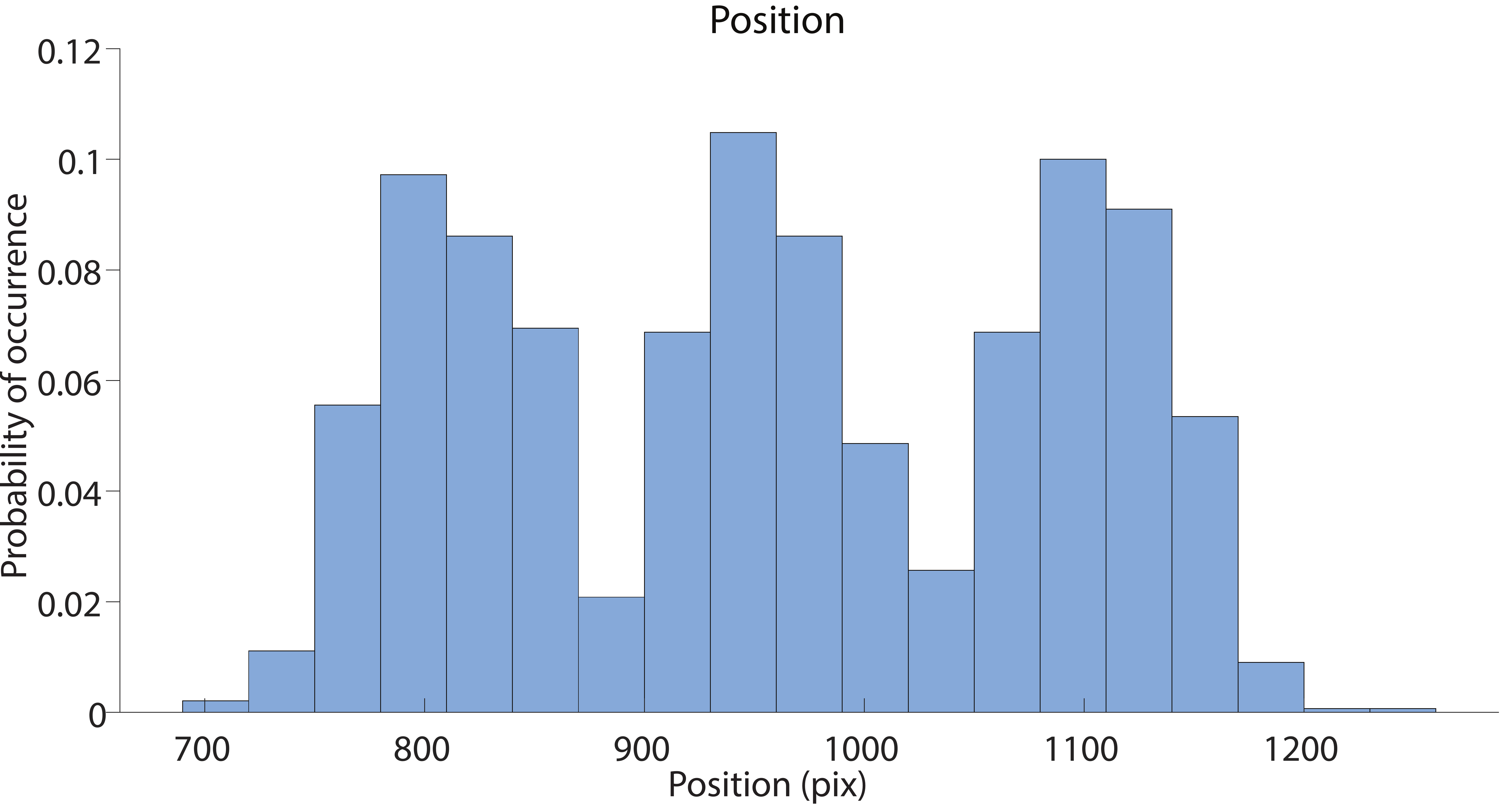}
       }\\
             \subfloat[\footnotesize Lum hist.\label{subfig-histo-luminosity}]{%
               \includegraphics[width=0.9\textwidth]{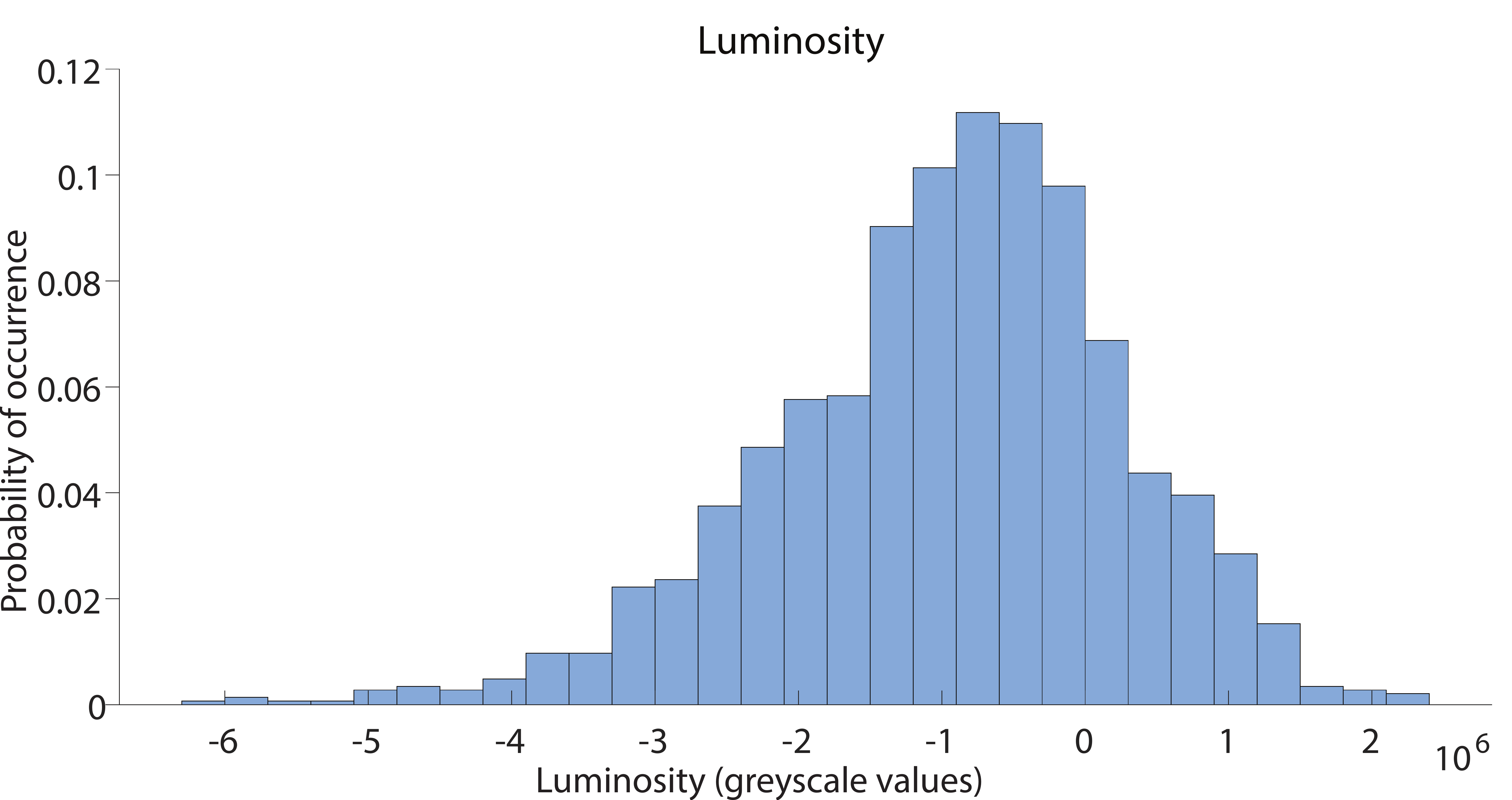}
             }  
          \end{minipage}%
        }\usebox{\bigleftbox}%
        \begin{minipage}[b][\ht\bigleftbox][s]{.6\textwidth}
        	\centering
             \subfloat[\footnotesize Pos\label{subfig-distance-position}]{%
       \includegraphics[width=0.32\textwidth]{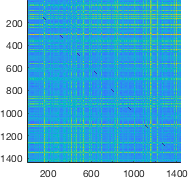}
     }
     \subfloat[\footnotesize Lum\label{subfig-distance-luminosity}]{%
       \includegraphics[width=0.32\textwidth]{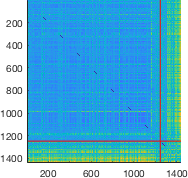}
     }
     \subfloat[\footnotesize Pos+Lum\label{subfig-distance-poslum}]{%
       \includegraphics[width=0.32\textwidth]{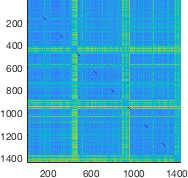}
     }\\
     \subfloat[\footnotesize Pos\label{subfig-1:tsne_pos}]{%
       \includegraphics[width=0.32\textwidth]{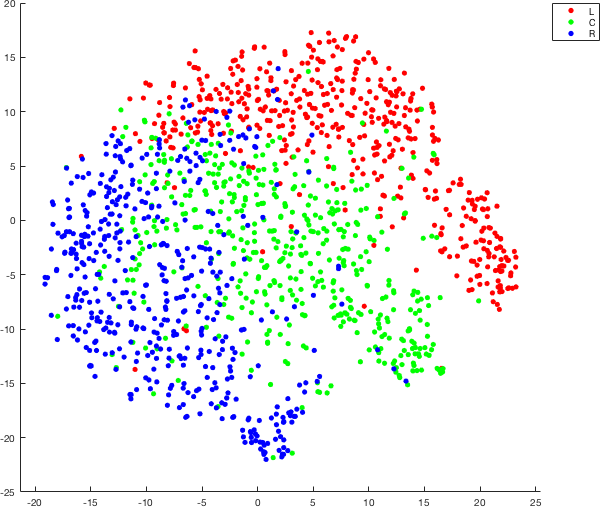}
     }
     \subfloat[\footnotesize Lum\label{subfig-2:tsne_lum}]{%
       \includegraphics[width=0.32\textwidth]{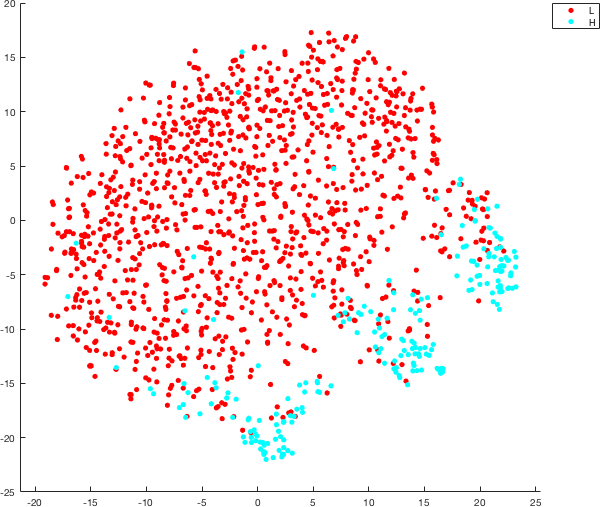}
     }
     \subfloat[\footnotesize Pos + Lum\label{subfig-3:tsne_poslum}]{%
       \includegraphics[width=0.32\textwidth]{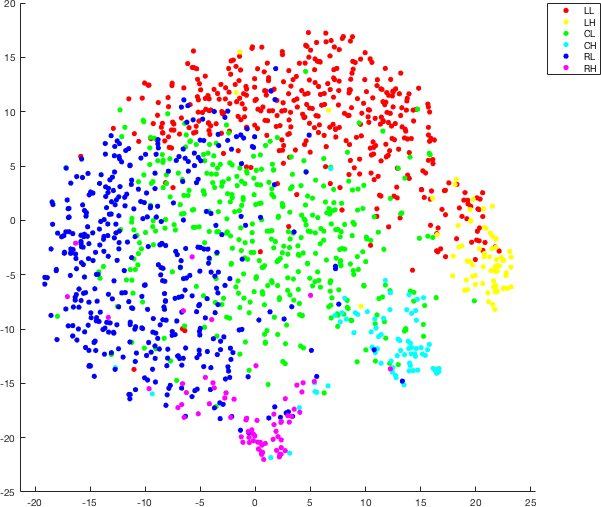}
     }
        \end{minipage}
        \vspace{-2mm}
        \caption{\footnotesize\textbf{Feature distribution and binning visualization.} Figures~\ref{subfig-histo-position}-~\ref{subfig-histo-luminosity} report position and luminosity histograms. Figures~\ref{subfig-distance-position}-~\ref{subfig-distance-luminosity} show distance matrices of neuronal population vectors ordered by position and luminosity. Figure~\ref{subfig-distance-poslum} shows the same distance matrix in which rows and columns are ordered according to the binning of the combined feature.  Figures~\ref{subfig-1:tsne_pos}-~\ref{subfig-3:tsne_poslum} report tSNE~\cite{vanDerMaatenHinton2008} 2-dimensional maps where point color refers to the feature binning.}
        \label{fig:ds_results}
\end{figure}
\subsection{Characterization of visual features}
To quantify the low- and mid-level features of the stimuli, we saved them as they were presented during the experiments and we extracted  the position of the center of mass of each stimulus and the total luminosity w.r.t. the background, defined as $L_{tot}=\sum(I(i,j)-128)$, where $I(i,j)$ is the matrix of pixel intensities in greyscale values.
As shown in Figure \ref{subfig-histo-position}, the distribution of the position of the stimuli along the $x$ axis was, as expected from the presentation protocol, trimodal, with the three peaks corresponding to the three main visual field positions used to show the stimuli during the experiment. This naturally leads to partition the set of stimuli in three classes, according to their position: left, right, or center. The luminosity instead shows a unimodal distribution that does not suggest a clear categorization; for this reason we visually inspected the distance matrix of the neuronal population vectors corresponding to each object, ordered according to the luminosity of the objects (see figure \ref{subfig-distance-luminosity}). We then set a threshold (red lines in figure \ref{subfig-distance-luminosity}) by hand at the point where the stimuli clearly separate.
The final distribution of samples (objects) per class is reported in Table \ref{tbl:classes}. 


\section{Characterize object representations through clustering}\label{sec:unsupervised_analysis}
First, we analyzed the space of neuronal responses obtained from V1 through unsupervised (clustering) methods. The aim was to assess how performing is the rat's neuronal embedding of the visual stimuli in terms of automatic grouping. The rationale is that stimuli having similar photometric characteristics (position and/or luminosity) should lie close to each other in the embedding, while being well separated from those having different properties. To check whether the neuronal mapping was meaningful in this regard, we tested different clustering algorithms, considering their best parameter setting under both internal and external indexes. 
The performances of each method have been stressed to their limit, so as to provide a guideline for future studies relying on similar methods.
To have an intuition of the complexity of this task we first visually inspected the distance matrices of all the stimuli per classes (see figure \ref{subfig-distance-position}-\ref{subfig-distance-poslum}). The distance matrix is a symmetric matrix of size $n \times n$ (where $n=1440$) containing the Euclidean distances of all the neuronal population vectors for each pair of stimuli. Then the matrix is sorted accordingly to the visual feature (position, luminosity and the binned position+luminosity) under exam.
As one can note, the matrix relative to the class position resembles a random matrix, and the three classes (left, center, right) are not clearly identifiable. Instead, the other two matrices, related to luminosity and the combination between position and luminosity, clearly report the two and six classes in which features are binned.

\vspace{-0.2cm}
\subsection{Experiments}
\vspace{-0.1cm}
    






\begin{table}[!t]
\begin{minipage}{.45\linewidth}
\begin{scriptsize}
\begin{tabular}{c l|cc|c}
& & \multicolumn{2}{c|}{Luminosity} & \multirow{2}{*}{Tot. Pos.}\\
& & Low lum. & High lum. & \multirow{2}{*}{}\\ \hline
\multicolumn{1}{c}{\multirow{3}{*}{Pos.}} & Right & 411 & 69 & 480 \\
\multicolumn{1}{c}{\multirow{3}{*}{}} & Center & 406 & 74 & 480 \\
\multicolumn{1}{c}{\multirow{3}{*}{}} & Left & 407 & 73 & 480 \\ \hline
\multicolumn{2}{c|}{Tot. Lum.} &  1224 & 216 & \textbf{1440} \\
\end{tabular}
\vspace{2mm}
\caption{\footnotesize Classes distributions.}
\label{tbl:classes}
\end{scriptsize}
\end{minipage}
\hfill\begin{minipage}{.5\linewidth}
\begin{center}
\begin{scriptsize}
    \begin{tabular}{c|c|c}
      Algorithm&K&SIL\\\hline\hline

      \rowcolor[HTML]{EFEFEF} DBSCAN eps=14.9 minPts=4&2&0.525\\

      DS $\sigma=504$&2&\textbf{0.565}\\\hline\hline

      \rowcolor[HTML]{EFEFEF} K-Means k=2&2&0.448\\

      K-Medoids k=2&2&0.394\\

    \end{tabular}
    \caption{\footnotesize Best Average Silhouette value for each compared clustering method and number of clusters extracted.}
    \label{tab:results_internal}
\end{scriptsize}
\end{center}
\end{minipage}
\vspace{-0.5cm}
\end{table}







    
    


\begin{figure}[b!]
\vspace{-0.5cm}
     \subfloat[\footnotesize DBSCAN\label{subfig-1:dbscan}]{%
       \includegraphics[width=0.24\textwidth]{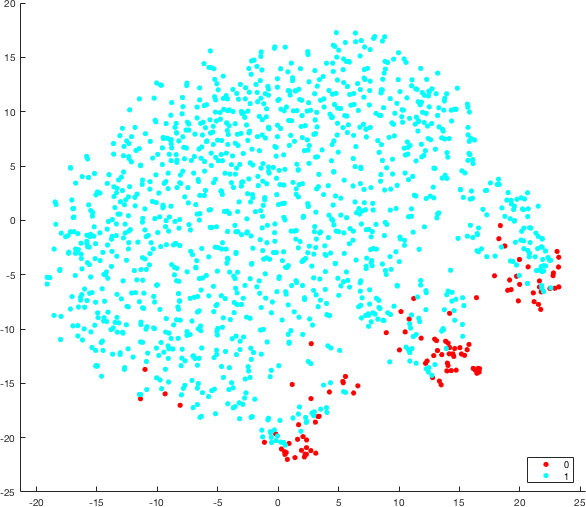}
     }
     \subfloat[\footnotesize DS\label{subfig-1:DS}]{%
       \includegraphics[width=0.24\textwidth]{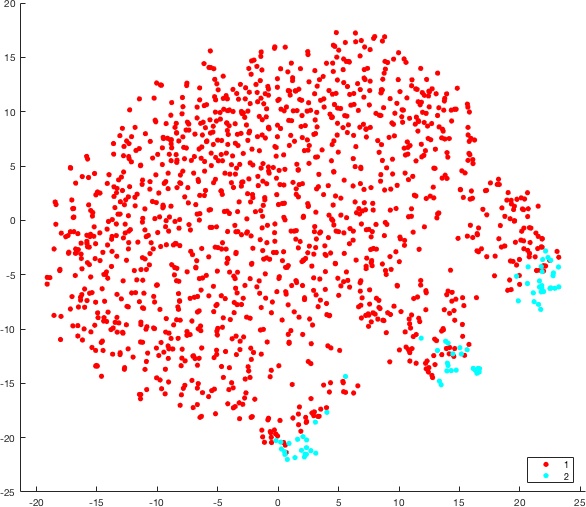}
     } 
     \subfloat[\footnotesize $k$-Means\label{subfig-1:kMeans}]{%
       \includegraphics[width=0.24\textwidth]{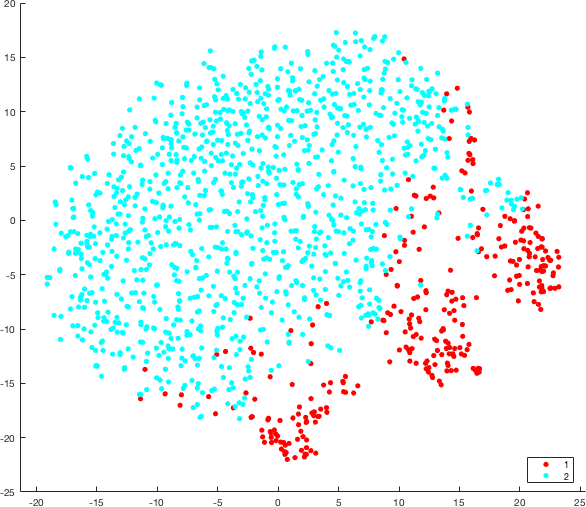}
     }
     \subfloat[\footnotesize $k$-Medoids\label{subfig-1:kMedoid}]{%
       \includegraphics[width=0.24\textwidth]{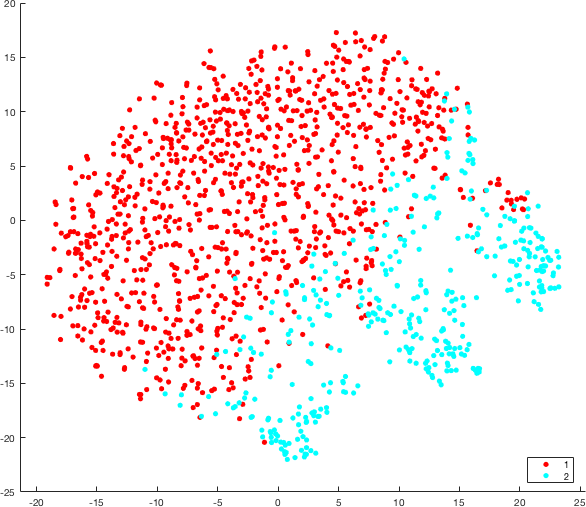}
     }
     \vspace{-1mm}
     \caption{\footnotesize \textbf{Best partition according to SIL.} Each point of the map is colored according to the cluster it belongs. Clusters are obtained with different methods through Silhouette maximization.}
     \label{fig:exp_unsup_internal}
\end{figure}
The experiments that have been carried out compared the performances of supervised clustering algorithms like $k$-means \cite{kmeans} and $k$-medoids\cite{kmedoid} (where the number of cluster is known in advance) and unsupervised techniques like \emph{DBSCAN}\cite{dbscan} and \emph{Dominant Set}\cite{pavan2007dominant}, where no a-priori information on the underlying structures is available.
We performed two different experiments considering on a first instance an internal criterion, the \emph{Silhouette} \cite{silhouette}, and later an external one, the \emph{Adjusted Rand Index} (ARI). The silhouette is a specific measure for each object and accounts for how well each object lies within its cluster and how well is separated from the others. In order to provide a global measure, the \emph{overall average silhouette width} (SIL) is taken \cite{silhouette}. The ARI is a measure that accounts for the agreement of two partitions, the predicted from a clustering method and the annotation: the higher its value, the better the algorithm has separated data. For all the clustering methods, the Euclidean metric has been used to compute distances/similarities.

\vspace{-0.5cm}
\subsubsection{Internal criterion}
For each method we searched in its parameters space the setting that maximizes the SIL. 
Maximizing the average silhouette means finding the parameters of a partitioning algorithm that separate and merge points in the best way possible provided their similarities or dissimilarities. In case a clustering method collapsed to an unwanted solution (one single cluster), we looked for the highest SIL value which separates the objects into at least two clusters with minimum density equal to the size of the less represented class of objects (in our case 69, see Table \ref{tbl:classes} for the classes distributions). We amend that this particular selection criterion is not completely fair, because it mixes some prior-information on the structure of the data with an internal index, but has the positive effect to find a reasonable solution for all the clustering algorithms at hand. The quantitative and qualitative results are reported in Table \ref{tab:results_internal} and in Figure \ref{fig:exp_unsup_internal} respectively. 
\vspace{-0.4cm}
\subsubsection{External criterion}
\begin{figure}[t!]
\footnotesize
     \subfloat[\footnotesize Best unsuper. method (DS $\sigma = 37$) for classes Pos and Pos + Lum, \label{subfig:bum_poslum_ds}]{%
       \includegraphics[width=0.22\textwidth]{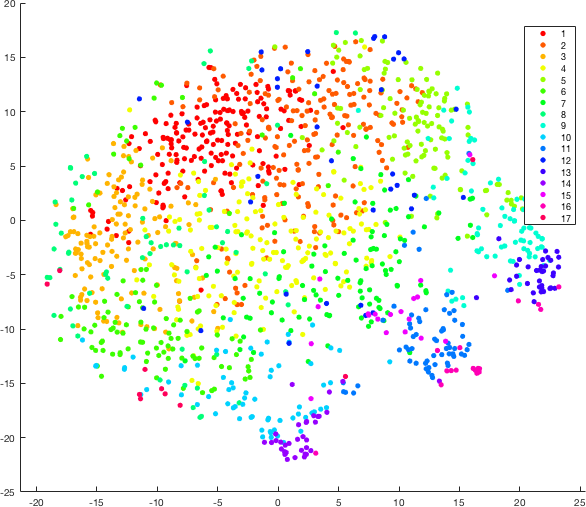}
       
     }\hfill
     \subfloat[\footnotesize Best unsuper. method (DS $\sigma=308$) for class Lum\label{subfig:bum_poslum_ds2}]{%
       \includegraphics[width=0.22\textwidth]{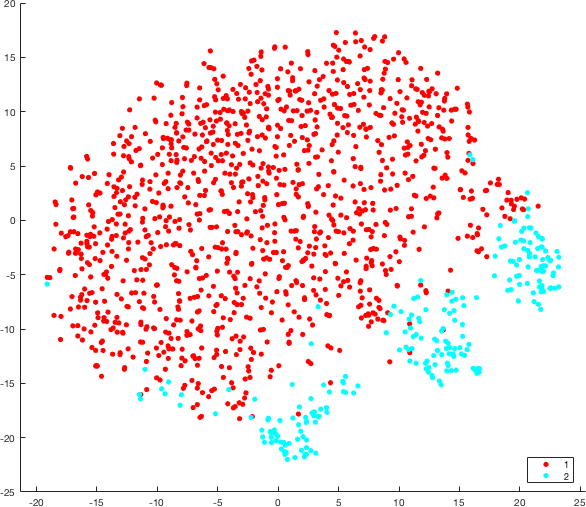}
     }\hfill
     \subfloat[\footnotesize Best super. method ($k$-means $k=6$) for classes Pos and Pos + Lum\label{subfig:bsm_poslum_km}]{%
       \includegraphics[width=0.22\textwidth]{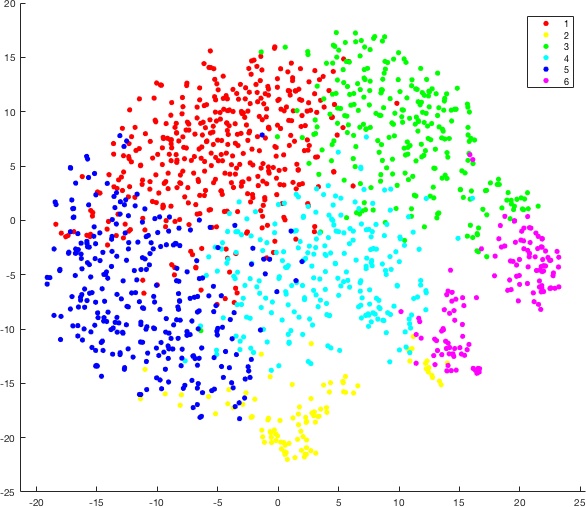}
     }\hfill
     \subfloat[\footnotesize Best super. method ($k$-means $k=2$) for class Lum\label{subfig:bsm_lum_km}]{%
       \includegraphics[width=0.22\textwidth]{images/k_2_tsne_cluster.png}
     }
     \caption{\footnotesize \textbf{Best partition according to ARI.} For each stimulus feature we selected the best partition produced by supervised ($k$-means and $k$-medoids) and unsupervised method (DS and DBSCAN) by maximizing ARI metric. Each point of the map is colored according to the cluster it belongs to.}
     \label{fig:exp_unsup_external}
\end{figure}

The test on the external criteria has been performed following the same schema as in the experiment on the internal measures, but instead of maximizing the SIL we looked for the parameters that maximize the ARI for each class of stimuli (see Table \ref{tbl:classes}). Other external indexes that are computed are the Adjusted Mutual Information (AMI) and the Purity (P). The AMI is similar to the ARI but quantifies the commonalities between two partitioning from an information-theoretic perspective. The purity index takes into account how the labels are organized inside of each cluster. 
We performed these experiments to understand which method has the potential of grouping as expected the different neuronal mappings with respect to the single classes. The quantitative and qualitative results are reported in Table \ref{tab:results_external} and in Figure \ref{fig:exp_unsup_external} respectively.


\begin{table}[t!]
\begin{scriptsize}
\begin{center}
    \begin{tabular}{c|cccc|cccc|cccc}
    \multirow{2}{*}{Alg.}&\multicolumn{4}{c|}{Position}&\multicolumn{4}{c|}{Luminosity}&\multicolumn{4}{c}{Position \& Luminosity}\\
    
    \multirow{2}{*}{}&K&AMI&P&ARI&K&AMI&P&ARI&K&AMI&P&ARI\\ \hline\hline

\rowcolor[HTML]{EFEFEF} DBSCAN&2&0.008&0.388&0.012&2&0.43&0.924&0.633&2&0.128&0.31&0.136\\
DS&17&0.179&\textbf{0.746}&0.121&2&\textbf{0.475}&\textbf{0.933}&\textbf{0.675}&17&0.268&\textbf{0.672}&0.171 \\ \hline\hline
\rowcolor[HTML]{EFEFEF} Kmeans&6&\textbf{0.255}&0.692&\textbf{0.233}&2&0.336&0.89&0.534&6&\textbf{0.378}&0.638&\textbf{0.299} \\
Kmedoids&3&0.189&0.557&0.179&2&0.257&0.85&0.415&6&0.313&0.543&0.21 \\

    \end{tabular}
    \vspace{1mm}
    \caption{\textbf{Best ARI value for each compared clustering algorithm and stimulus classes.} The best clustering algorithm for each feature and external index is highlighted in bold.}
    \label{tab:results_external}
\end{center}
\end{scriptsize}
\vspace{-1cm}
\end{table}

\subsection{Results}
In general the values of SIL, ARI and AMI are quite low indicating that the task is very complex, data is very noisy and visual stimuli are not perfectly mapped to the neuronal response. This can be easily seen in Figures \ref{subfig-1:tsne_pos}-\ref{subfig-3:tsne_poslum},
where the tSNE~\cite{vanDerMaatenHinton2008} projection shows a mixing of classes, in particular on the upper part of the projection for the position (Figure \ref{subfig-1:tsne_pos}) and position+luminosity (see Figure \ref{subfig-3:tsne_poslum}). The luminosity classes, instead, are well separated from each other. From both perspectives (internal and external indexes)
the embedding found using the responses from V1
was sufficiently able to group the different classes. 

Considering the internal index (see Table \ref{tab:results_internal}), the method that best performs was the DS, outperforming also the supervised methods like $k$-means and $k$-medoids. The motivations are due to the fact that DS only depends
on the similarity matrix, which is not the case for the other techniques that also rely
on other assumptions (like the number of clusters or global densities). 
Regarding the results for the external criterion (see Table \ref{tab:results_external}) the best unsupervised clustering method is the DS among all features. 
The top purity is reached by the DS and this can be explained by the higher number of clusters that are generated. In terms of supervised clustering, the best performing method is $k$-means in all the considered metrics.
These results suggest us that, in case of absent a-priori information on the number of clusters, the DS method can be considered as a more-than-valid alternative to standard approaches (like DBSCAN). Furthermore, knowledge on the number of clusters can be fruitfully used by supervised clustering algorithms (like $k$-means).
\section{Inferring object properties with supervised learning}\label{sec:supervised_analysis}

As seen in section \ref{sec:unsupervised_analysis}, the analysis of the neuronal embedding was meaningful under
different criteria to analyze how the space is partitioned. This lead to a second set of experiments in terms of discrimination power of the features extracted from the V1 area. We considered separately the three classes of photometric characteristics \emph{position}, \emph{luminosity} and \emph{position+luminosity} and carried out several tests by training and testing standard classifiers (Linear/Kernel SVM\footnote{\tiny Software at \url{https://www.csie.ntu.edu.tw/~cjlin/libsvm/}}, Error Correcting Output Code Linear SVM \footnote{\tiny Software at \url{https://www.mathworks.com/help/stats/fitcecoc.html}} \cite{dietterich1994solving} and k-NN) on the V1 embedding to confirm its discrimination capability. The rationale is that similar visual stimuli will lie in close proximity and vice-versa different ones will be located far away. With this assumption, a classifier should be able to find a boundary to discriminate between the classes. 
\subsection{Experiments \& Results}

\begin{table}[t!]
\begin{scriptsize}
\begin{center}
\begin{tabular}{l|cccc|cccc|cccc}
\multicolumn{1}{c|}{}                      & \multicolumn{4}{c|}{Position}                                                                                                                              & \multicolumn{4}{c|}{Luminosity}                                                                                                                            & \multicolumn{4}{c}{Position + Luminosity}        \\
\multicolumn{1}{c|}{\multirow{-2}{*}{Alg}} & ACC                                               & AUC                                               & mF1 & MF1                                                 & ACC                                               & AUC                                               & mF1 & MF1                                                 & ACC            & AUC            & mF1 & MF1             \\ 
\hline
\hline
\cline{2-10} 
\rowcolor[HTML]{EFEFEF} 
Lin SVM                                    & 0.906                                             & 0.961                                             & 0.859 &  0.859                                            & 0.913                                             & 0.922                                             & 0.913  & 0.818                                           & 0.756          & 0.500          & 0.268  &  0.181     \\
Ker SVM                                    & \textbf{0.942}                                    & \textbf{0.979}                                    & \textbf{0.912}   & \textbf{0.912}                                 & \textbf{0.939}                                    & \textbf{0.970}                                    & \textbf{0.939}   & \textbf{0.877}                            & 0.758          & 0.536          & 0.275  &  0.181      \\
\rowcolor[HTML]{EFEFEF} 
ECOC LSVM                                  & 0.908                                             & 0.963                                             & 0.862    & 0.862                                         & 0.912                                             & 0.922                                             & 0.912     & 0.813                                        & \textbf{0.937} & \textbf{0.966} & \textbf{0.810} & \textbf{0.778}\\
\rowcolor[HTML]{FFFFFF} 
$k$-NN ($k=9$)                                       & 0.913 & 0.960 & 0.870 & 0.870 & 0.931 & 0.937 & 0.931 & 0.855 & 0.935          & 0.944  & 0.805  & 0.763  
\end{tabular}
\vspace{1mm}
\caption{\footnotesize \textbf{Classifier performances.}}
\label{tbl:classifier_results}
\end{center}
\end{scriptsize}
\vspace{-0.8cm}
\end{table}
Considering a class of visual stimuli (see Table \ref{tbl:classes} for a details on classes) we performed a 5-fold cross validation to find the best parameter setting of each classifier. The folds have been created in a stratified way ensuring that each class is represented with the same proportion of the dataset. The training and testing have been performed randomly generating 10 different splits of the data and consequently averaging the performances. To evaluate the performances we used four indexes that are common in classification tasks: \emph{accuracy} (ACC), \emph{average area under the ROC curve} (AUC), the \emph{micro F-measure} (mF1) and the \emph{macro F-measure} (MF1) \cite{Manning:2008:IIR:1394399}. The results are reported in Table \ref{tbl:classifier_results}.
It is evident how the neuronal responses can be used successfully to classify visual stimuli; in fact, we achieved a very high ACC, AUC and (in general) F-score. As expected, due to the class imbalance (see Table \ref{tbl:classes}), the MF1 is a bit lower than the mF1. This is particularly evident for the luminosity and position+luminosity classes, in which a strong imbalance (the larger class is $\simeq6$ times bigger w.r.t. the smaller one) is reported. Regarding the class position and the class luminosity, the best performing method was the Kernel SVM followed by the $k$-NN. It is worth to note that the simple $k$-NN is the second best choice for all the three classes; this gives us an indication of the difficulties in finding a linear separator, hence on the non linear separability of the space. This motivates also the fact that the Linear SVM performs poorly w.r.t. the Kernel SVM. Furthermore, in the case of position+luminosity the best performing method was the ECOC L-SVM followed by the $k$-NN. This is explained by the fact that, in that particular case, we increased the number of classes from 2-3 to 6, needing more hyperplanes to separate them. The ECOC is based on an ensemble of Linear SVMs trained in one-vs-one mode which creates all the possible intersecting hyperplanes w.r.t. the classes. For this reason it outperforms the other classifiers, while being not so far from the performances of the Linear SVM for the Position and for the Luminosity, both cases having fewer classes. Concerning the stability of the results we reported a maximum mean standard deviation of $\simeq1\%$ considering all the 10 runs. 
\section{Conclusions}\label{sec:conclusion}
In this paper, we investigated how visual stimuli are mapped into the representational space of V1 neurons focusing on two low-level properties (luminosity and position within the visual field). We thus quantified the extent to which these properties were accurately represented in the V1 population space, using supervised and unsupervised learning methods. We found that, indeed, both luminosity and position and their combination are naturally mapped in the V1 representation, and that these features can be accurately extracted using pattern classifiers. Among the clustering methods, DS showed the greatest accuracy at inferring the structure of the representation. Among the classifiers, the SVM with nonlinear kernel achieved the highest accuracy. In both cases, this testifies of the complexity of the representation and of the not complete linear discriminability of the data.

As future work, we will try different distance functions and will test whether other higher-level visual features, e.g. orientation, are encoded. Moreover, the same data processing pipeline will be applied to higher-oder visual areas, e.g. LM-LI-LL, to understand the differences with V1.

\section*{Acknowledge}
This work was supported by a European Research Council Consolidator Grant (DZ, project n. 616803-LEARN2SEE).

\bibliographystyle{splncs04}

\end{document}